\documentclass{article}

\PassOptionsToPackage{square,numbers}{natbib}
\usepackage[preprint]{neurips_2023}

\usepackage[utf8]{inputenc} %
\usepackage[T1]{fontenc}    %
\usepackage[backref=page]{hyperref}       %
\usepackage{url}            %
\usepackage{booktabs}       %
\usepackage{amsfonts}       %
\usepackage{nicefrac}       %
\usepackage{microtype}      %
\usepackage{xcolor}         %
\usepackage{amsmath}
\usepackage[ruled,vlined]{algorithm2e}
\usepackage{multirow}
\usepackage{booktabs}
\usepackage{graphicx}
\usepackage{float}
\usepackage{subcaption}
\usepackage{wrapfig}
\usepackage{nicefrac}
\usepackage{enumitem}

\title{Mix-ME: Quality-Diversity for Multi-Agent Learning}

\newcommand{\ucl}{1}
\newcommand{\imperial}{2}

\author{%
    \textbf{Garðar Ingvarsson}$^\ucl{}$
    \textbf{Mikayel Samvelyan}$^{\ucl{}}$
    \textbf{Bryan Lim}$^{\imperial{}}$
    \textbf{Manon Flageat}$^{\imperial{}}$\\
    \textbf{Antoine Cully}$^{\imperial{}}$
    \textbf{Tim Rockt{\"a}schel}$^\ucl{}$\\[0.25em]
    $^\ucl{}$University College London  ~${}^\imperial{}$Imperial College London\\[0.25em]
    \texttt{gardarjuto@gmail.com}
}

\begin{document}

\maketitle

\begin{abstract}
  In many real-world systems, such as adaptive robotics, achieving a single, optimised solution may be insufficient. Instead, a diverse set of high-performing solutions is often required to adapt to varying contexts and requirements. This is the realm of Quality-Diversity (QD), which aims to discover a collection of high-performing solutions, each with their own unique characteristics. QD methods have recently seen success in many domains, including robotics, where they have been used to discover damage-adaptive locomotion controllers. However, most existing work has focused on single-agent settings, despite many tasks of interest being multi-agent. To this end, we introduce Mix-ME, a novel multi-agent variant of the popular MAP-Elites algorithm that forms new solutions using a crossover-like operator by mixing together agents from different teams. We evaluate the proposed methods on a variety of partially observable continuous control tasks. Our evaluation shows that these multi-agent variants obtained by Mix-ME not only compete with single-agent baselines but also often outperform them in multi-agent settings under partial observability.
\end{abstract}

\section{Introduction}
The conventional paradigm of optimisation has largely focused on finding a single, optimal solution that performs exceptionally well on a given problem. However, for many real-world tasks, there is need for solutions that exhibit varied behaviour across different contexts or dimensions. In such scenarios, the concept of quality-diversity~\citep[QD,][]{lehman2011abandoning, Cully2018Quality} comes into play.

QD methods aim to discover a diverse set of high-performing solutions that span different dimensions of a problem space. Unlike traditional optimisation methods that converge to a single optimal solution, QD methods produce a population of solutions that are both high-quality and diverse. This is particularly useful in problems where a single ``best'' solution is either not sufficient or not meaningful. For example, in robotic locomotion, there is need for strategies that adapt to malfunctions. For a robot with a damaged limb, the optimal movement pattern would differ significantly from its undamaged state. Therefore, discovering a collection of diverse gaits ensures robustness against unforeseen damages~\citep{cully2015robots, colas2020scaling}.

The realm of multi-agent systems (MAS) presents a fertile ground for the application of QD methods. In many real-world systems, multiple agents interact in a shared environment to achieve a common goal. These systems are often partially observable, meaning that each agent has a limited view of the full state of the environment. For instance, in robotic control, there might be latency or bandwidth constraints~\citep{peng2020facmac, ong2010planning} that limit the amount of information that can be shared between different parts of the robot. In such cases, each body part needs to act intelligently based on its own partial information~\citep{takadama2003ARL}.

Despite clear benefits, QD has not been extensively applied to multi-agent learning. Most mainstream works in the field of multi-agent systems rely on traditional optimisation methods that do not capture the essence of diversity across solutions. Furthermore, those works that do train for diversity are usually based on mutual information, making it hard to specify the type of diversity induced. Applications of MAP-Elites~\citep{mouret2015illuminating}, a popular QD algorithm, to multi-agent problems have been limited to either rule-based agents~\citep{canaan2019diverse} or environments providing dense
agent-specific rewards~\citep{dixit2022balancing}, presenting a significant gap in the literature.

This paper addresses this gap by exploring how MAP-Elites can
be extended to cooperative multi-agent problems, specifically in partially observable continuous control tasks. 
We propose \textit{Mix-ME}, a novel extension of the MAP-Elites algorithm to the multi-agent setting. Mix-ME maintains a set of solutions and progressively refines them by generating new ones through random mutation and a crossover mechanism that mixes together agents from different teams. 

We rigorously compare Mix-ME to a naive multi-agent baseline and against single-agent policies through empirical evaluation, including a sensitivity analysis on policy network size as well as generalisation experiments. This comparative analysis provides insight into the strengths and weaknesses of each approach, adding to our understanding of how open-ended learning methods can be effectively applied in various multi-agent settings~\citep{samvelyan2023maestro}.

\vspace{-0.2cm}
\section{Related Work}
\paragraph{Single-Agent QD}
Quality-Diversity (QD) methods have been successfully applied to a variety of single-agent
continuous control tasks. Much of this stems from the work of \citet{cully2015robots}, who
introduced the MAP-Elites algorithm and demonstrated its effectiveness for damage adaptation
in robotic locomotion. Since then, MAP-Elites has seen widespread use in the robotics community,
with QDax, a recent JAX-based library of QD algorithms by \citet{lim2022accelerated}, enabling massive speedup on acceleration hardware. They also show that MAP-Elites can be parallelised by batching multiple grid updates in a single step, without sacrificing performance.
This has brought training times down from days to minutes, making
works such as this possible on a reasonable time scale.
More recently, \citet{chalumeau2023neuroevolution} have shown that MAP-Elites and its derivatives
are competitive with deep RL diversity algorithms in terms of fitness and skill discovery, despite
the former being simpler and less sample-efficient. The authors tested their methods on a variety
of continuous control tasks, including the unidirectional Ant, Walker2d and HalfCheetah tasks,
which we also use in our experiments. 
The work of \citet{allard2022hierarchical} bears some resemblance to ours, as they also decompose the robot's movement into separate limbs movements. Using MAP-Elites, they compute a hierarchical structure of grids, where each grid is responsible for a different
level of abstraction. This parallels our approach of decomposing the robot into multiple
controllers. However, they use a centralised algorithm to determine the next action and the
individual leg controllers do not have policies of their own, but execute a sequence
of predefined movements.

\paragraph{Multi-Agent QD}
Despite the recent success of QD methods in single-agent settings, there is limited work on applying them to multi-agent problems. Some work has been done on ad-hoc teamwork and zero-shot coordination (ZSC) in the game of Hanabi: \citet{canaan2019diverse} use MAP-Elites with self-play to train a collection of agents, however, their agents are rule-based; ADVERSITY by \citet{cui2023adversarial} is a RL method to produce diverse teams of agents for turn-based games with public actions; and TrajeDi by \citet{lupu2021trajectory} produces diverse and robust policies for ZSC, based on a generalised Jensen-Shannon Divergence. Ridge Rider, proposed by \citet{parker2020ridge}, is a novel method for exploring the loss landscape by following the eigenvectors of the Hessian. They achieve diverse solutions effective for ZSC in a simple coordination game. Another work, by \citet{li2021celebrating}, achieves diversity between agents by maximising mutual information between agents' identities and trajectories, improving on previous Google Research Football \citep{kurach2020google} and StarCraft II \citep{samvelyan2019starcraft} benchmarks. Unsupervised environment design (UED) is yet another approach for achieving diversity, as shown by \citet{samvelyan2023maestro}, who use UED to design a curriculum for training a population of diverse agents for robustness in zero-sum games.
Finally, a more QD-like algorithm, coupled with PPO, is used by \citet{dixit2022balancing} to train a team of agents in a cooperative 2D exploration game. Their algorithm shows promising results, however, it requires dense
agent-specific rewards, which are not always available in real-world scenarios.

\section{Background}

\subsection{Quality-Diversity}

Quality-diversity~\citep[QD,][]{lehman2011abandoning, Cully2018Quality} is a paradigm of evolutionary computation where the aim is to discover 
a diverse set of high-performing solutions that span different dimensions of a problem space.
Whereas traditional optimisation methods aim to find a single solution $x\in\mathcal{X}$ that
maximises an objective
function $\mathtt{fitness}:\mathcal{X}\mapsto\mathbb{R}$, QD methods aim to find a collection of
solutions $X\subset\mathcal{X}$ where each solution $x\in X$ is high-performing in a different
way. This diversity is defined in terms of a solution's behaviour descriptor (or feature vector)
$\mathtt{behaviour\_descriptor}:\mathcal{X}\mapsto\mathcal{B}$, that maps the solution to a vector of features
that describe its behaviour, attributes or characteristics. 
The behaviour descriptor is a parameterisation of what kind of diversity we are interested in 
and is hand-crafted based on the characteristics of the problem domain.

\begin{wrapfigure}[22]{R}{0.5\textwidth}
\vspace{-0.5cm}
\begin{algorithm}[H]
  \caption{MAP-Elites Algorithm}
  \DontPrintSemicolon
  \SetKwProg{Init}{Initialise}{:}{}
  \label{alg:mapelites}
  \KwIn{Initial number of solutions $N_\text{initial solutions}$, number of iterations $N_\text{iterations}$}
  \KwOut{A grid $X$ of high-performing solutions}
  \Init{}{
    Create $D$-dimensional grid of solutions $X$ and fitnesses $F$\;
    Populate the grid with $N_\text{initial solutions}$ random solutions.\;
  }
  \For{$i=1$ \KwTo $N_\text{iterations}$}{
    $x \gets \mathtt{sample\_solution}(X)$\;
    $x' \gets \mathtt{mutate}(x)$\;
    $f \gets \mathtt{fitness}(x')$\;
    $\mathbf{b}' \gets \mathtt{behaviour}(x')$\;
    \If{$f > F[\mathbf{b}']$ \textbf{or} $X[\mathbf{b}']$ is empty}{
      $X[\mathbf{b}'] \gets x'$\;
      $F[\mathbf{b}'] \gets f$\;
    }
  }
  \end{algorithm}
\end{wrapfigure}
\paragraph{MAP-Elites}
MAP-Elites~\citep{mouret2015illuminating} is one of the fundamental QD algorithms and underlies 
most of current research in the field.
In its simplest form, MAP-Elites discretises the behaviour space into an initially empty grid $X$
of cells with the same dimensionality as the behaviour descriptor. Each cell in the grid
can hold one solution, called an \emph{elite}. In the case of two solutions having the same 
behaviour descriptor, the algorithm only keeps the one with the higher fitness. 
Before starting the main loop, the algorithm populates the grid with $N_\mathrm{initial\ solutions}$ random solutions.
Then, each iteration proceeds by sampling a random solution $x$ from the grid $X$,
randomly mutating it to produce an offspring $x'$, evaluating $x'$ and computing its behaviour descriptor
$\mathbf{b}'$. Using $\mathbf{b}'$, the algorithm looks up the relevant cell in the grid. 
If $x'$ has higher fitness than the current elite in cell $\mathbf{b}'$, the elite is replaced
with $x'$. This process is repeated for $N_\mathrm{iterations}$ iterations, gradually filling the grid
with high-performing solutions. The full algorithm is shown in Algorithm~\ref{alg:mapelites}.

A big advantage of MAP-Elites is that it is highly parallelisable. In practice, the algorithm is
implemented by running multiple instances of the main loop in parallel. 
This allows for massive parallelisation, which is a big driver of the algorithm's success~\cite{lim2022accelerated}.
This counterbalances the fact that QD approaches usually require a large number of iterations to reach good solutions.

\subsection{Cooperative Multi-Agent Learning}

In this work, we consider partially observable cooperative multi-agent problems defined using DecPOMDP~\citep{oliehoek2016concise}.
   Dec-POMDP is a 7-tuple $(\mathcal{S}, \{ \mathcal{A}^{(i)} \}_{i=1}^N, \mathcal{P}, r, \{ \mathcal{Z}^{(i)} \}_{i=1}^N, \mathcal{O}, \gamma)$, where $\mathcal{S}$ is the set of possible states of the environment; $\mathcal{A}^{(i)}$ is the set of actions available to agent $i$; ${\mathcal{P}: \mathcal{S} \times \mathcal{A}^{(1)} \times \dots \times \mathcal{A}^{(N)} \times \mathcal{S} \mapsto [0,1]}$ is the transition probability function, 
     where $\mathcal{P}(s'|s,a^{(1)},\dots,a^{(N)})$ is the probability of transitioning to state $s'$ after the agents simultaneously take actions $a^{(1)},\dots,a^{(N)}$ in state $s$; $r: \mathcal{S} \times \mathcal{A}^{(1)} \times \dots \times \mathcal{A}^{(N)} \mapsto \mathbb{R}$ is the expected reward $r=\mathbb{E}\left[R \mid s, a^{(1)},\dots,a^{(N)}\right]$
    received after the agents take actions $a^{(1)},\dots,a^{(N)}$ in state $s$; $\mathcal{Z}^{(i)}$ is the set of observations available to agent $i$; $\mathcal{O}: \mathcal{S} \times \mathcal{Z}^{(1)} \times \dots \times \mathcal{Z}^{(N)} \mapsto [0,1]$ is the observation probability function, 
    where ${\mathcal{O}(z^{(1)},\dots,z^{(N)} \mid s)}$ is the probability of observing $z_1,\dots,z_N$ after transitioning to state $s$; $\gamma \in [0,1]$ is the discount factor for trading off immediate and future rewards.
At each time step $t$, every agent $i$ receives a partial observation $z_t^{(i)}$ of the environment state $s_t$, and then they all independently, but simultaneously, select actions $a_t^{(1)}, \dots, a_t^{(N)}$ based on their own policies.
The environment then transitions to a new state $s_{t+1}$ according to the transition probability function $\mathcal{P}$, and the agents receive a joint reward $r_{t+1}$ according to the reward function $r$.
The goal of the agents is to learn a joint policy $\pi=(\pi_1,\dots,\pi_N)$ that maximises the expected return ${J(\pi)=\mathbb{E}\left[\sum_{t=0}^\infty \gamma^tr_{t+1} \mid \pi\right]}$. Since agents do not individually have access to the full environment state, they must learn to collaborate in order to achieve the goal.

\section{Methods}
In this section, we present the design of the proposed multi-agent QD approaches.
All of the methods described below are based on the same core MAP-Elites algorithm, 
which is described in Algorithm~\ref{alg:mapelites}. However, we assume two main changes to the definitions:
\begin{enumerate}
    \item The parameter space $\mathcal{X}$ is now a set of $N$ parameter spaces $\mathcal{X}_1, \dots, \mathcal{X}_N$, one for each agent.
    \item Solutions in the grid are now tuples $(x_1, \dots, x_N)$, where $x_i$ is the solution for agent $i$.
\end{enumerate}
The first change is necessary since agents can have different action and observation spaces, and therefore need to be sampled from
different parameter spaces. The second change is needed to allow the algorithm to keep track of each individual agent policy.

\subsection{Naive Multi-Agent MAP-Elites}
\begin{figure}
  \centering
  \includegraphics[width=\textwidth]{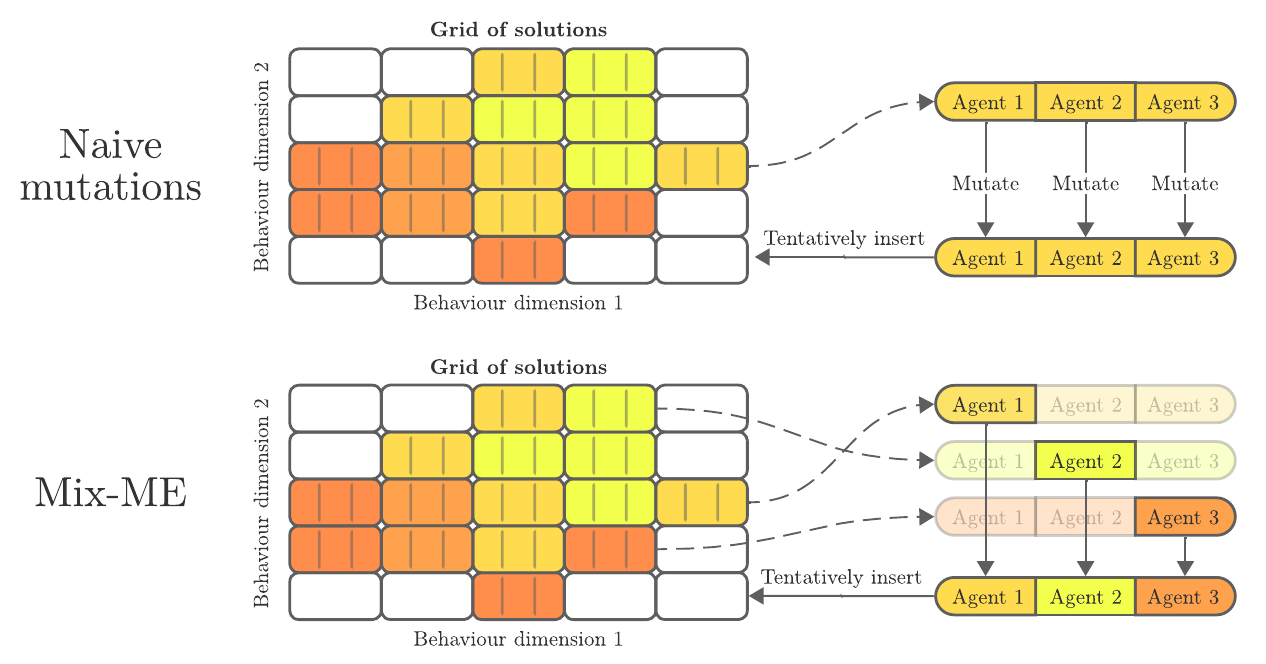}
  \caption{Graphical illustration of the difference between naive mutations and the team-mixing operation proposed in Mix-ME.}
  \label{fig:naive_mame}
\end{figure}

The most straightforward way to train groups of cooperative agents with MAP-Elites is to treat 
the group as a single unit, and use the single-agent variation of MAP-Elites.
In this approach, a new offspring is created by sampling a random solution $\left(x_1, \dots, x_N\right)$ 
from the grid and then mutating each of the agents' policies $x_i$ to produce a new policy $x_i'$.
The resulting team of agents $\left(x_1', \dots, x_N'\right)$ is then evaluated and assigned to the grid.
The algorithm is illustrated in Figure~\ref{fig:naive_mame}\textbf{(top)}. The mutations we use are polynomial mutation~\citep{kalyanmoy1999niched} and isoline variation ~\citep[Iso-LineBB,][]{vassiliades2018discovering}.

This approach is simple and easy to implement, but its restrictive sampling strategy can
limit its potential. The fact that every agent in the offspring is derived from the same 
parent solution means that the algorithm is not able to combine policies from different parents.
This might lose out on some potential benefits of co-adaptation between agents.

\subsection{Mix-ME}
One relaxation of the baseline approach is to allow the agents in the offspring to be derived 
from different parents. In a multi-agent system, different agents might have specialised roles 
that require different capabilities or expertise. The motivation for this approach is that 
during training, we might have multiple solutions in the grid containing agents that are 
proficient in different roles. By allowing agents in an offspring to inherit policies from different 
parents, the algorithm can combine experts from different teams and therefore promote the 
co-adaptation of agents with complementary roles.
An analogy to this approach is the formation of sports teams, where the coach might select a 
strong goalkeeper from one team, a strong striker from another team, and so on.

We thus introduce the \textit{Agent Mixing MAP-Elites} (Mix-ME), a novel multi-agent QD approach that performs mix-and-matching of individual agents between distinct groups within the grid.
In Mix-ME, in addition to using naive mutations on raw parameters, it includes a team-crossover operator. This operator creates a new offspring by
sampling $N$ random solution tuples 
$\left(x_1^{(1)}, \dots, x_N^{(1)}\right), \dots, \left(x_1^{(N)}, \dots, x_N^{(N)}\right)$
with replacement from the grid. The offspring is then created by taking the 1st agent from 
the 1st tuple, the 2nd agent from the 2nd tuple, and so on. The resulting team of agents
$\left({x_1^{(1)}}, \dots, {x_N^{(N)}}\right)$ is then evaluated and assigned to the grid.
This operator is illustrated
in Figure~\ref{fig:naive_mame}\textbf{(bottom)}. 

Since massive parallelism in MAP-Elites is achieved by producing new solutions in batches, in
each iteration, we split the batch evenly across operators. Thus, with a batch size of $N_\text{batch}$, $\nicefrac{N_\text{batch}}{3}$ new solutions would be formed using polynomial mutation, $\nicefrac{N_\text{batch}}{3}$ with isoline variation, and $\nicefrac{N_\text{batch}}{3}$ using team-crossover. The purpose of the conventional mutation operators is to optimise the weights, resulting in local hill-climbing behaviour, while the purpose of the team-crossover operator is to promote co-adaptation of agents with complementary roles.

\section{Experimental Setup}
In this section, we explain the motivation, design and setup for our experiments, as well as describing the training environments.
The main questions we seek to answer are:
\begin{enumerate}
  \item How do the proposed algorithms compare against each other and against the single-agent baseline in 
  terms of performance, diversity and generalisation capability?
  \item Do specific traits of the environment affect the performance of the proposed algorithms?
  \item How does changing the size of the policy networks impact the performance of the different MAP-Elites methods?
\end{enumerate}

For details on our experimental implementation and hyperparameter settings, please refer to Appendices~\ref{appendix:implementation} and~\ref{appendix:hyperparameters}.

\paragraph{Environments}
\label{sec:environments}
To evaluate the proposed methods, we extend five existing single-agent continuous control environments in the Brax physics engine~\citep{brax2021github} to support multiple agents. These environments are the multi-agent parallels of the single-agent Mujoco environments~\citep{todorov2012mujoco} and were first introduced by \citet{peng2020facmac}.
Previous implementations, however, have not natively supported JAX~\citep{jax2018github}, and their parallelisability has been limited. Our implementation uses pure JAX and is highly parallelisable, allowing for massive speedup on acceleration hardware. Moreover, it is compatible with the QDax library~\citep{lim2022accelerated, chalumeau2023qdax} which we base our QD algorithms on.

\begin{figure}
  \centering
  \begin{subfigure}[b]{0.15\textwidth}
    \centering
    \includegraphics[height=\linewidth]{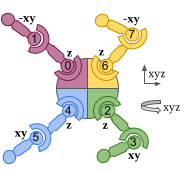}
    \caption{Ant}
  \end{subfigure}\hspace{0.5cm}
  \begin{subfigure}[b]{0.15\textwidth}
    \centering
    \includegraphics[height=\linewidth]{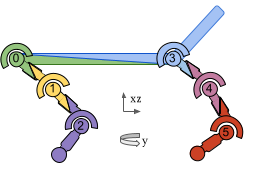}
    \caption{HalfCheetah}
  \end{subfigure}\hspace{1cm}
  \begin{subfigure}[b]{0.15\textwidth}
    \centering
    \includegraphics[height=\linewidth]{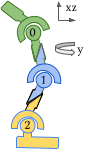}
    \caption{Hopper}
  \end{subfigure}
  \begin{subfigure}[b]{0.15\textwidth}
    \centering
    \includegraphics[height=\linewidth]{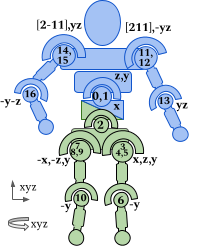}
    \caption{Humanoid}
  \end{subfigure}\hspace{0.5cm}
  \begin{subfigure}[b]{0.15\textwidth}
    \centering
    \includegraphics[height=\linewidth]{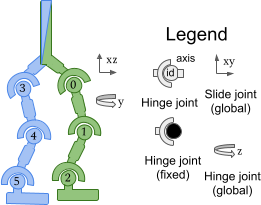}
    \caption{Walker2D}
  \end{subfigure}
  \caption{Illustration of the robot factorisations. The colours represent the different agents. Image sourced from \citet{peng2020facmac}.}
  \label{fig:environments}
\end{figure}

The environments adapt the single-agent MuJoCo tasks to multi-agent use with the concept of factored robots. In this paradigm, the robot is partitioned into multiple components, 
each controlled by an individual agent. Figure~\ref{fig:environments} illustrates the factorisation for each environment. These agents have partial observability of the global 
state and act based on local information. 
They must then collaboratively control the robot to accomplish the task.
The environment specifics are described in Table~\ref{tab:environments}, and 
in more detail in Appendix~\ref{appendix:environments}. 

\begin{table}[ht]
  \caption{Summary of the environments used in our experiments.}
  \label{tab:environments}
  \centering
  \begin{tabular}{lllllll}
    \toprule
    \textbf{Environment} & \textbf{Agents} & \multicolumn{2}{l}{\textbf{Observation Space}} & \multicolumn{2}{l}{\textbf{Action Space}} \\
    \cmidrule(r){3-4} \cmidrule(r){5-6}
    & & Single-Agent & Multi-Agent & Single-Agent & Multi-Agent \\
    \midrule\midrule
    Ant & 4 & 28 & (18 each) & 8 & (2 each) \\
    HalfCheetah & 6 & 18 & (9, 9, 8, 8, 9, 8) & 6 & (1 each) \\
    Hopper & 3 & 11 & (8, 9, 8) & 3 & (1 each) \\
    Humanoid & 2 & 376 & (248, 176) & 17 & (9, 8) \\
    Walker2D & 2 & 22 & (17 each) & 6 & (3 each) \\
    \bottomrule
  \end{tabular}
\end{table}

The behaviour descriptor we use for all environments is the average time that each foot of the
robot is in contact with the ground during an episode, represented by 
\begin{equation}
  \mathbf{b} = \frac{1}{T} \sum_{t=1}^T \begin{pmatrix}
    \mathbb{I}[\text{foot 1 touches ground}] \\
    \mathbb{I}[\text{foot 2 touches ground}] \\
    \vdots \\
    \mathbb{I}[\text{foot N touches ground}]
  \end{pmatrix}
\end{equation}
where $T$ is the length of the episode, and $\mathbb{I}$ is the indicator function.
This behaviour descriptor is simple but effective for capturing various gaits of the robot.
For example, a hopping gait would have a low value for all feet, while a walking gait would have
a higher value. It has also been used in previous studies~\citep{cully2015robots}
to allow robots to recover from mechanical damage.

\paragraph{Evaluation Metrics}
We use the following three metrics when comparing the performances of baselines.
Firstly, the \emph{maximum fitness} $f_{\text{max}} = \max_{f\in F} f$, i.e. the fitness of the best performing solution in the grid at the end of training, where fitness refers to the total reward received during an episode.
Secondly, the \emph{coverage} $C = \frac{\text{number of cells containing a solution}}{\text{total number of cells}}$, representing the proportion of the behaviour space that solutions have been found for. The coverage is a measure of the diversity of the solution grid.
Thirdly, we measure the \emph{QD score}, $QD = \sum_{f\in F} f$, the sum of the fitnesses of all solutions in the grid at the end of training. This score summarises both performance and diversity of the solution grid and is the main metric we use to compare MAP-Elites methods against each other.

\paragraph{Generalisation Experiments}
To assess the generalisation capabilities of our proposed algorithms~\citep{mahajan2022generalization}, we follow the experimental 
procedure outlined in a previous paper by \citet{chalumeau2023neuroevolution}. This procedure 
employs a few-shot adaptation approach in modified environments, where pre-computed policies are evaluated
without retraining. We explore two distinct settings: gravity update and leg dysfunction. 
In the gravity update scenario, the gravity constant is 
modified by multiplying it with a coefficient over a specified range. In the leg dysfunction setting, 
we alter the input-to-torque coefficients of a single leg across a range. 

Initially, each baseline is 
trained for 1,000 iterations in a standard environment. Then, we conduct 100 evaluations for 
each solution in the grid using the modified environments, calculating the median fitness for 
each solution. The maximum of these median fitness values is then reported. To ensure robustness 
and reliability, we report the results of the experiments across 10 different seed values.

\section{Results and Discussion}

\subsection{Comparison of Multi-Agent MAP-Elites Methods}
\label{sec:comparison_of_multi_agent_map_elites_methods}

\begin{figure}[ht]
  \centering
  \includegraphics[width=\textwidth]{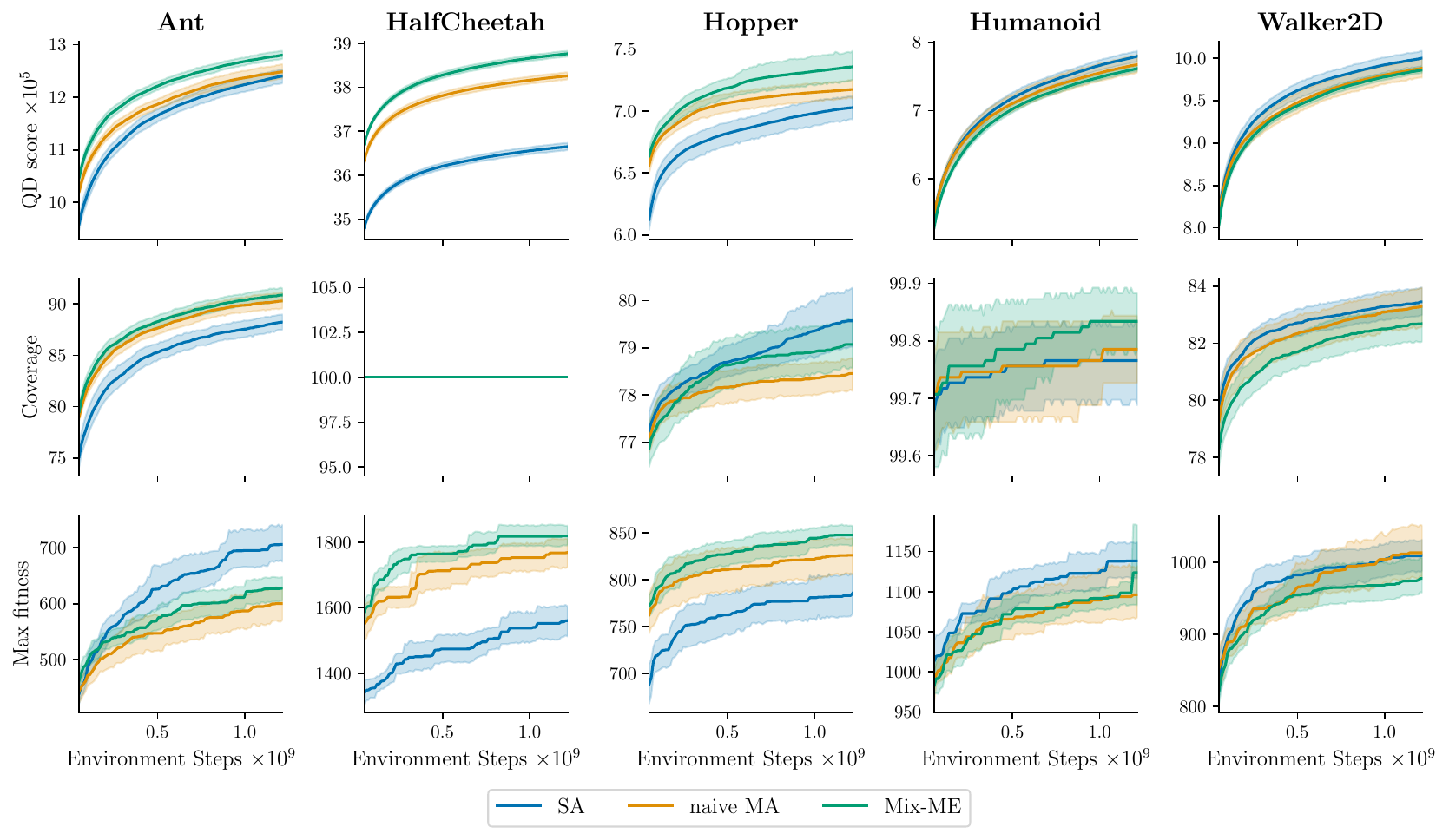}
  \caption{Learning curves of the multi-agent (Naive MA \& Mix-ME) and single agent (SA) MAP-Elites methods.
  The shaded regions represent the standard deviation across 10 runs. On the x-axis, we show the total number of environment steps
  taken by the algorithm, which is equal to the number of iterations multiplied by the number of offspring per iteration.}
  \label{fig:mamapelites}
\end{figure}

\paragraph{Performance and Diversity}
We first compare the performance of the naive multi-agent MAP-Elites baseline, Mix-ME, 
and the single-agent baseline. 
Figure~\ref{fig:mamapelites} illustrates the learning curves for each of the environments. %
We can observe several interesting trends. 

First, we see that across environments with more than 2 agents 
(Ant, HalfCheetah, Hopper), Mix-ME consistently outperforms the naive multi-agent baseline 
on every metric. On the other hand, in the Walker2d environment, we see the opposite trend, where the naive
baseline outperforms Mix-ME, only slightly in terms of QD score, but significantly in terms
of coverage and maximum fitness. This suggests that mixing elites builds better 
teams by exploiting diversity in the solution grid, but loses its effectiveness
when the number of agents is small.
\begin{wrapfigure}{r}{0.46\textwidth}
  \centering
  \includegraphics[width=0.46\textwidth]{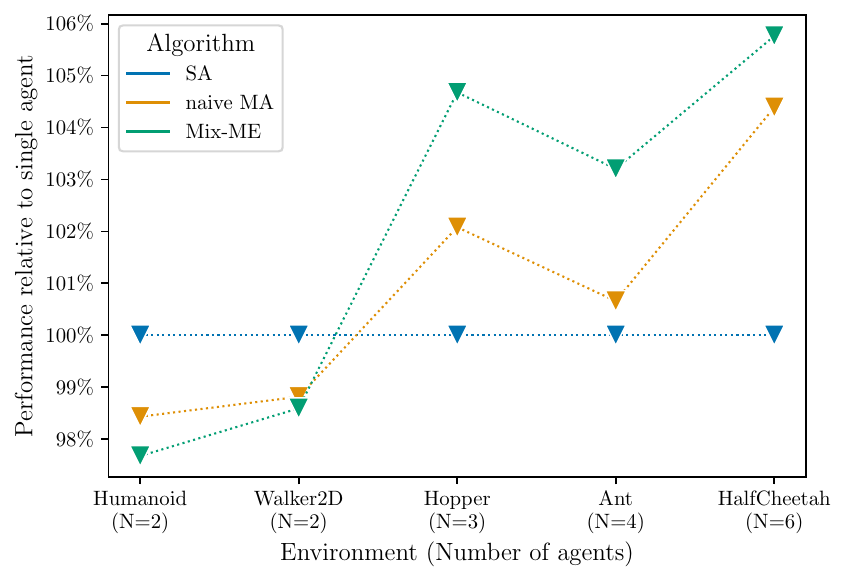}
  \caption{Performance (QD score) at the end of training, as a percentage relative to the single-agent baseline, ordered by number of agents.}
  \vspace{-0.5cm}
  \label{fig:performance_difference}
\end{wrapfigure}

Another interesting observation is that in terms of QD score, both multi-agent methods outperform the single-agent
baseline in environments with more than 2 agents. In fact, if we look at the score as a function of the number of agents,
we see that the performance gap roughly increases with the number of agents, as shown in
Figure~\ref{fig:performance_difference}. 
An important caveat, however, is that the multi-agent
methods have the same policy network architecture per agent as the single-agent baseline, which means that
the total number of parameters in the multi-agent methods is comparably larger. 
We explore the effect of policy network size in Section~\ref{subsec:policy_network_size} and show that
this fact alone does not explain the difference in performance. 
Therefore, the multi-agent methods are indeed able to learn a higher-performing solution grid.

Apart from performance, in Appendix~\ref{appendix:behaviour_descriptor_grid} we also show the resulting solution
grids for each environment. We do not see any obvious differences between the grids of different methods, 
except for the Ant environment, where the solution grid for the multi-agent methods seems 
to be more uniform than the single-agent baseline.
This is consistent with Figure~\ref{fig:mamapelites}, which shows higher QD-scores but lower maximum fitness for the multi-agent
methods than the single-agent baseline.  Partial observability
might be a factor here, since the agents only have access to local information, and therefore might not be able
to learn as high-performing policies as the single-agent baseline.

\begin{figure}[ht]
  \centering
  \includegraphics[width=\textwidth]{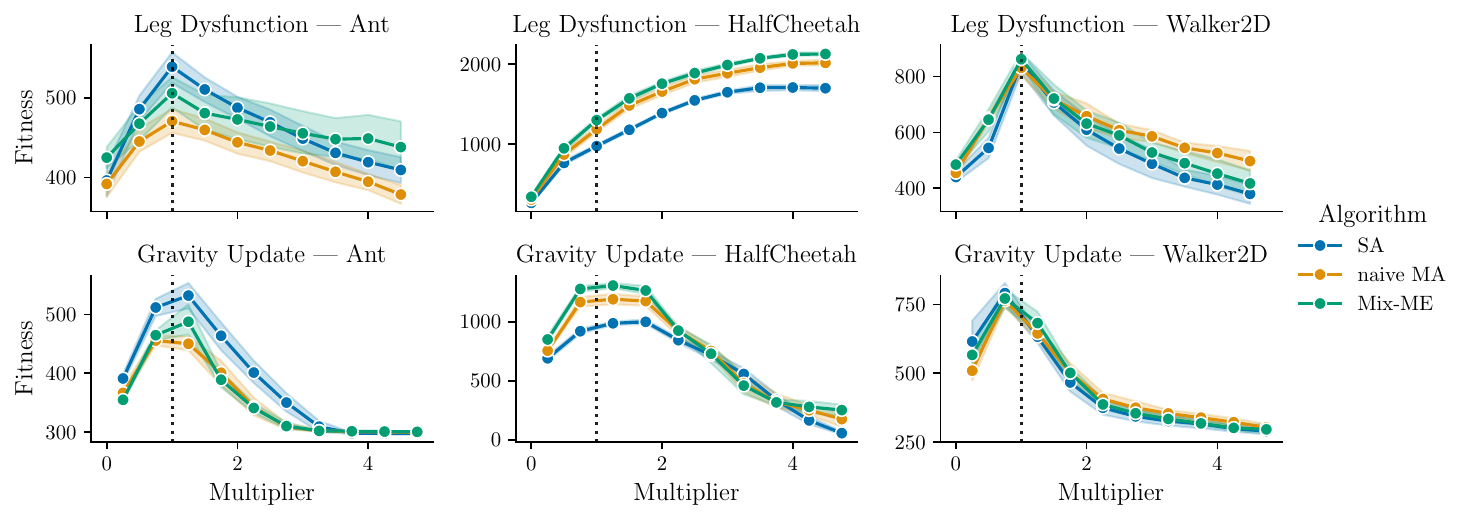}
  \caption{Generalisation results. The x-axis shows the multiplier applied to the gravity constant in the gravity update 
  scenario, and the coefficient applied to the input-to-torque coefficients in the leg dysfunction scenario. The y-axis shows the median fitness of the best solution in the grid.}
  \label{fig:generalisation}
\end{figure}

\paragraph{Generalisation}
We also evaluate the generalisation capabilities of the different methods in the 
leg dysfunction and  gravity update
scenarios. The results are shown in Figure~\ref{fig:generalisation} and show similar
trends to the results in the previous section. We see that Mix-ME
generalises better than the naive multi-agent baseline in environments with more than 2 agents,
but worse in the Walker2d environment. 

We also see that in the leg dysfunction scenario,
the performance of the single-agent baseline drops significantly faster than the multi-agent methods as the test environment diverges from the training environment. In the gravity update scenario, the results
are mixed and highly dependent on the environment.
A key difference between the two scenarios is that leg dysfunction is exactly the 
failure mode that the behaviour descriptor is designed to capture; the descriptor
is a parameterisation of each leg's contact time with the ground, and therefore
the resulting grid of solutions should contain solutions that are diverse in terms
of individual leg usage. On the other hand, it is not obvious how this kind of diversity
would be useful in the gravity update scenario.

\noindent
These results have a straightforward interpretation: since multi-agent methods
learn higher-performing and more diverse solutions in environments with many agents, if the 
behaviour descriptor is well-aligned with the task, naturally the multi-agent methods 
will be able to learn more robust solutions that are less sensitive to changes in 
the environment, compared to their single-agent counterpart.

\subsection{Effect of Policy Network Size}
\label{subsec:policy_network_size}
When comparing single-agent and multi-agent baselines, one subtle caveat is that each individual agent's policy network in the multi-agent methods has the same architecture as the single-agent baseline has for controlling the entire robot. As a result, the total number of parameters
in the multi-agent methods is $N_\text{agents}$ times larger than the single-agent baseline.
To address this issue, we conduct an experiment where we vary the size of the policy networks
in each method, and observe how the performance scales. Note that we performed hyperparameter
tuning for each policy network size separately to ensure optimal learning.

\begin{figure}[ht]
  \centering
  \includegraphics[width=\textwidth]{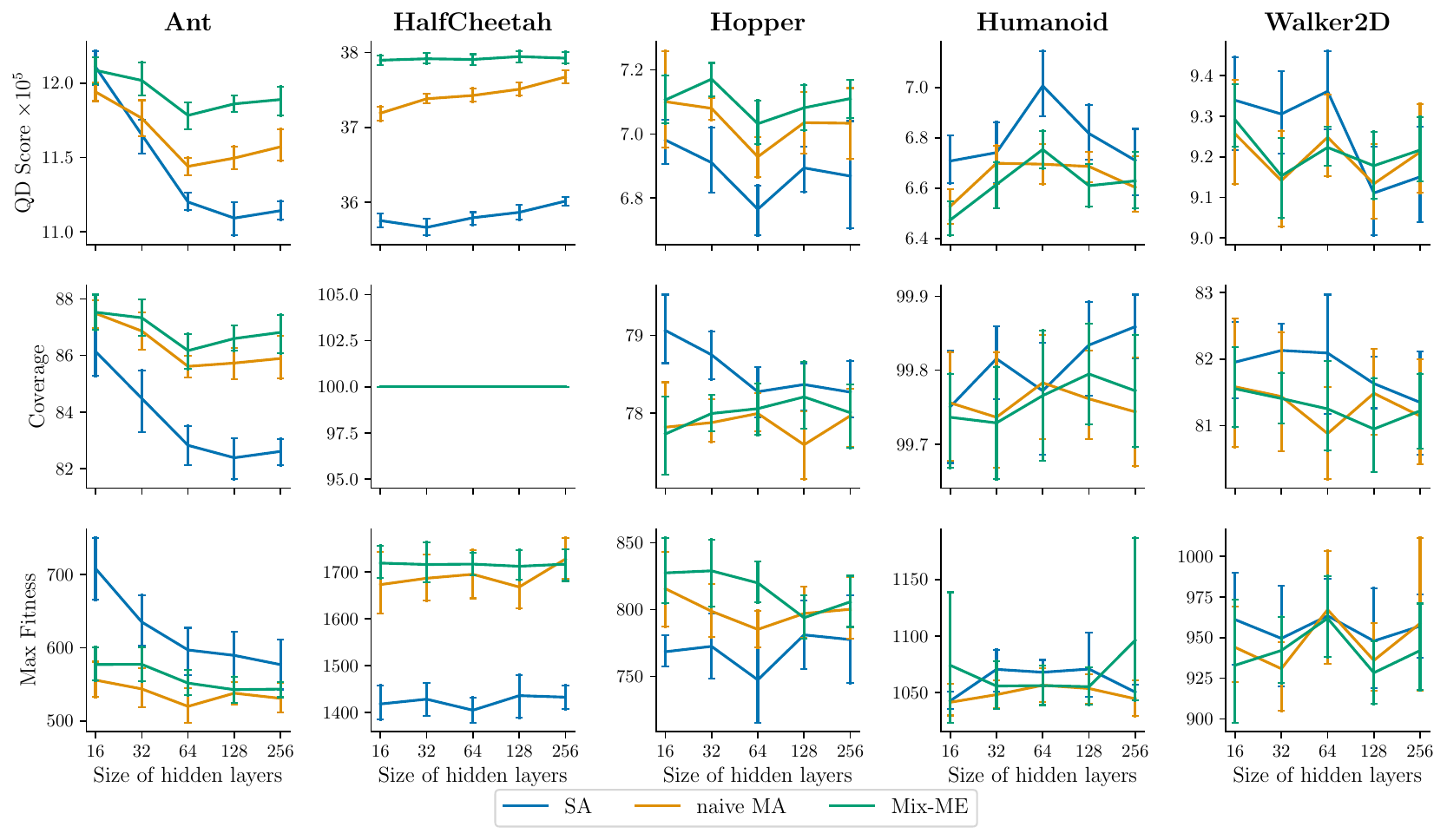}
  \caption{Effect of policy network size on performance of the different methods. The x-axis shows the number of units in each of the two hidden layers. 
  95\% confidence intervals are shown as error bars. Note that we have reduced the batch size here to 1024 to allow for larger networks, meaning absolute performance is not comparable to previous sections.}
  \label{fig:qd_score_policy_size}
\end{figure}

Figure~\ref{fig:qd_score_policy_size} shows the results of this experiment. We can see that
in none of the environments does increasing the policy network size of the single-agent baseline
result in comparatively better performance than the multi-agent methods with smaller policy networks.
In other words, we don not gain much by increasing the policy network size of the single-agent baseline.
In fact, in most environments, the performance either drops or stays the same with increasing policy network size.

This, in combination with the results from Section~\ref{sec:comparison_of_multi_agent_map_elites_methods},
 means that the good performance of the multi-agent methods cannot simply be attributed
to the larger number of parameters. Instead, it suggests that there must be some benefit to
learning a decentralised policy, even though this imposes partial observability on each agent.

\section{Conclusion and Future Work}
\label{chap:conclusion}
This work sets out to bridge the gap between QD and cooperative multi-agent learning. It was motivated by the observation that many real-world continuous control tasks are inherently partially observable and multi-agent, and that often, we are interested in inducing diversity in the solutions to these tasks, yielding a set of high-quality solutions, that are robust to damage and to changes in the environment.

To this end, we proposed Mix-ME, a new multi-agent variant of the MAP-Elites algorithm, which adds a team-crossover operation to form new solutions. We presented a comprehensive set of experiments that compare the performance of our proposed method against a naive multi-agent extension and against a single-agent baseline. These experiments revealed that Mix-ME shows superior performance and generalisation capabilities, and that this performance gap increases with the number of agents.
We also showed that in many-agent environments, decentralised control policies trained using Mix-ME outperform single-agent policies trained using normal MAP-Elites, even under partial observability.

There are numerous avenues for future work. First, benchmarking Mix-ME on different environments would be a good way to further validate the results. This paper only includes continuous control environments with a relatively small number of agents, and it would be interesting to see how the methods scale to environments with more agents. Environments with discrete action spaces, such as grid-worlds, would also be beneficial to explore.
Another avenue to explore is multi-agent extensions of more sophisticated MAP-Elites variants, such as Policy gradient assisted MAP-Elites~\citep[PGA-ME,][]{nilsson2021policy}, which employs first-order optimisation techniques. This could potentially lead to better scaling and performance, making MAP-Elites methods more competitive with policy gradient methods in terms of maximum fitness.
Our proposed methods extend easily to these variants, and therefore could be a good starting point for future work.

This paper has shown that MAP-Elites methods are a promising approach to inducing diversity in multi-agent learning. We hope that this work will inspire further research in this direction, and that it will help to bridge the gap between QD and cooperative multi-agent learning.

\medskip

\bibliography{neurips_2023}

\newpage
\appendix

\section{Appendix}

\subsection{Environments}
\label{appendix:environments}
\paragraph{Ant}
The Ant environment~\citep{schulman2018highdimensional} is a 3-dimensional 4-legged robot with 8 rotors. 
We factorise it into 4 agents, each controlling the two joints on one leg. 
The agents observe the angle and angular velocity of the local leg joints and immediately adjacent joints, as well as the
global position and velocity of the robot central body.

The goal is to make the Ant walk forward as fast as possible, while minimising energy consumption and
external contact forces. All agents receive a shared reward at each time step, defined as
\begin{equation}
  r := r_{\text{survive}} + r_{\text{forward}} - r_{\text{ctrl}} - r_{\text{contact cost}}
\end{equation}
where $r_{\text{survive}}$ is a constant reward for surviving, 
$r_{\text{forward}}$ is the forward velocity of the robot,
$r_{\text{ctrl}}$ is a penalty for large control inputs, and
$r_{\text{contact cost}}$ is a penalty for external contact forces.

\paragraph{HalfCheetah}
The HalfCheetah environment is a 2-dimensional 2-legged robot with 6 rotors. 
We factorise it into 6 agents, each controlling one joint. The agents observe the angle and 
angular velocity of their assigned joint and immediately adjacent joints, 
as well as the global position and velocity of the 
tip of the robot. 

The goal is to make the HalfCheetah run forward as fast as possible, while minimising energy consumption.
All agents receive a shared reward at each time step, defined as
\begin{equation}
  r := r_{\text{forward}} - r_{\text{ctrl}}
\end{equation}
where the individual reward components are the same as in the Ant environment.

\paragraph{Hopper}
The Hopper environment is a 2-dimensional 1-legged robot with 3 rotors.
We factorise it into 3 agents, each controlling one joint. The agents observe the angle and
angular velocity of their assigned joint and immediately adjacent joints, as well as the global position and velocity of the
top of the robot.

The goal is to hop forward as fast as possible, while minimising energy consumption.
All agents receive a shared reward at each time step, defined as
\begin{equation}
  r := r_{\text{survive}} + r_{\text{forward}} - r_{\text{ctrl}}
\end{equation}
where the individual reward components are the same as in the Ant environment.

\paragraph{Humanoid}
The Humanoid environment is a 3-dimensional 2-legged robot with 20 rotors, designed to resemble a human.
We factorise it into 2 agents, one controlling the upper body and the other controlling the lower body.
The agents observe the angle and angular velocity of their assigned joints and immediately adjacent joints, as well as the global position and velocity of the
humanoid's torso.

The goal is to make the Humanoid walk forward as fast as possible, while minimising energy consumption.
All agents receive a shared reward at each time step, defined as
\begin{equation}
  r := r_{\text{survive}} + r_{\text{forward}} - r_{\text{ctrl}}
\end{equation}
where the individual reward components are the same as in the Ant environment.

\subsection{Implementation Details}
\label{appendix:implementation}
In each of our experiments, we perform 10 runs with different random seeds and 
report the mean and standard deviation of the results.
Each run consists of 1000 iterations of the MAP-Elites algorithm, where each iteration 
produces 4096 offspring. We evaluate offspring in parallel for 300 timesteps on a single GPU for each job.
We use GPUs of types NVIDIA GTX 1080 Ti, RTX 2080 Ti, Tesla P100, V100, and A100.

\subsection{Hyperparameters}
\label{appendix:hyperparameters}
In order to ensure a fair comparison between the different methods, we tuned mutation hyperparameters 
for each combination of environment and policy network size. 
The hyperparameters were tuned by running a grid search over a range of values for each 
hyperparameter, and selecting the combination that yielded the highest QD score averaged over 3 seeds. 
These optimal hyperparameters were then used for all experiments. We used a fully connected multi-layer perceptron
with 2 hidden layers of 64 units each, save for the policy network sensitivity analysis where the hidden layer size was modified.

\begin{table}[H]
  \caption{Search space for MAP-Elites mutation hyperparameters.}
  \label{tab:hyperparameters}
  \centering
  \begin{tabular}{ll}
    \toprule
    \textbf{Hyperparameter} & \textbf{Search space} \\
    \midrule\midrule
    $\sigma_\text{iso}$ & $\{0.0001, 0.001, 0.01, 0.1, 1.0\}$ \\
    $\sigma_\text{line}$ & $\{0.0001, 0.001, 0.01, 0.1, 1.0\}$ \\
    $\eta$ & $\{4, 8, 16, 32, 64, 128, 256\}$ \\
    \bottomrule
  \end{tabular}
\end{table}

\begin{table}[H]
  \caption{Optimal MAP-Elites hyperparameters for each environment and policy network size.}
  \label{tab:optimal_hyperparameters}
  \centering
  \begin{tabular}{lllll}
    \toprule
    \textbf{Environment} & \textbf{Policy network hidden layer size} & $\sigma_\text{iso}$ & $\sigma_\text{line}$ & $\eta$ \\
    \midrule\midrule
    \multirow{5}{*}{Ant} & 16 & 0.001 & 1.0 & 32 \\
    & 32 & 0.001 & 1.0 & 64 \\
    & 64 & 0.001 & 1.0 & 128 \\
    & 128 & 0.001 & 1.0 & 128 \\
    & 256 & 0.001 & 1.0 & 128 \\
    \midrule
    \multirow{5}{*}{HalfCheetah} & 16 & 0.01 & 0.1 & 128 \\
    & 32 & 0.001 & 0.1 & 128 \\
    & 64 & 0.001 & 0.1 & 128 \\
    & 128 & 0.001 & 0.1 & 128 \\
    & 256 & 0.001 & 0.1 & 128 \\
    \midrule
    \multirow{5}{*}{Hopper} & 16 & 0.001 & 0.1 & 8 \\
    & 32 & 0.001 & 0.1 & 16 \\
    & 64 & 0.001 & 0.1 & 16 \\
    & 128 & 0.001 & 0.1 & 64 \\
    & 256 & 0.001 & 0.1 & 128 \\
    \midrule
    \multirow{5}{*}{Humanoid} & 16 & 0.001 & 1.0 & 32 \\
    & 32 & 0.001 & 1.0 & 64 \\
    & 64 & 0.001 & 1.0 & 128 \\
    & 128 & 0.001 & 1.0 & 128 \\
    & 256 & 0.001 & 1.0 & 128 \\
    \midrule
    \multirow{5}{*}{Walker2d} & 16 & 0.001 & 0.1 & 4 \\
    & 32 & 0.001 & 0.1 & 4 \\
    & 64 & 0.01 & 0.1 & 8 \\
    & 128 & 0.001 & 0.1 & 8 \\
    & 256 & 0.01 & 0.01 & 8 \\
    \bottomrule
  \end{tabular}
\end{table}

\newpage
\subsection{MAP-Elites Behaviour Descriptor Grid}
\label{appendix:behaviour_descriptor_grid}

\begin{figure}[htb]
  \centering
  \includegraphics[width=0.9\textwidth,trim={0 7cm 0 6cm},clip]{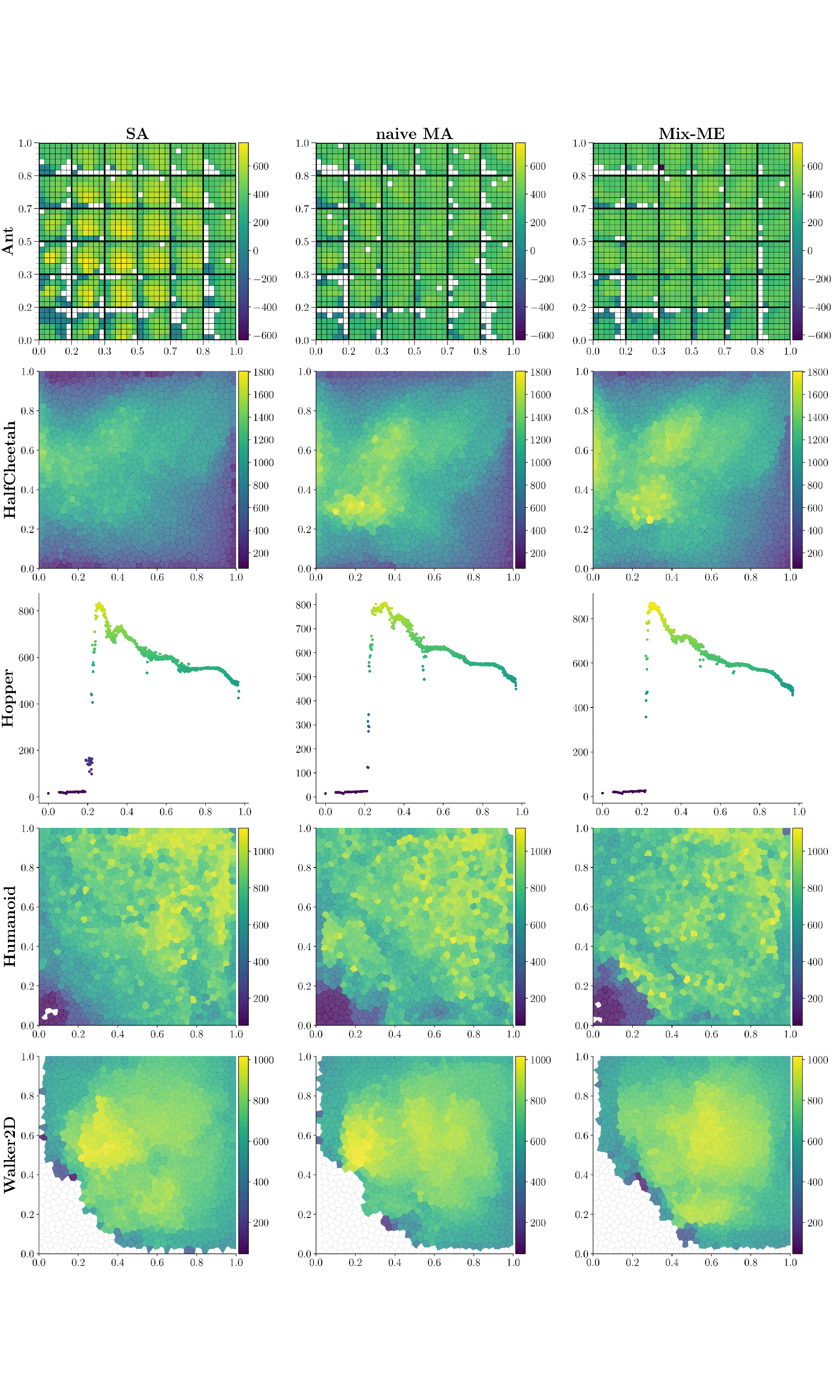}
  \caption{Visualisation of the solution grids produced by the different multi-agent MAP-Elites methods, broken down by environment. 
  The visualisation for the 4-dimensional descriptor in the Ant environment is projected into 2D as is done in \citet{cully2015robots}.}
  \label{fig:behaviour_grid}
\end{figure}

\end{document}